# Unseen Harm: Measuring Cross-Script Safety Inconsistency with "Missed-in-Urdu" Scores in LLM Hate Speech Detection

**Fawzia Z Kara-Isitt**
Brunel University of London, UK
Fuzzy.Kara-Isitt@brunel.ac.uk

**Sonal Khosla**
Hasso Plattner Institute Potsdam, Germany
sonal.khosla@hpi.de

**Stephen Swift**
Brunel University of London, UK
Stephen.Swift@brunel.ac.uk

## Abstract

Urdu, the world's tenth most spoken language with 246 million speakers, remains almost entirely absent from mainstream LLM safety evaluation and nine years of WOAH proceedings. To investigate whether this absence has measurable consequences for content moderation reliability, five large language models, GPT-4o, Claude Sonnet 4.5, Gemini 2.5 Flash, Qwen-2.5, and Llama-3.1, were tested across six datasets spanning Nastaliq Urdu, Roman Urdu, English, and code-switched UrduEnglish. Across the five Urdu-script datasets, label instability between original-script and English-translation classification ranged from 15.9% (Gemini 2.5 Flash) to 31.6% (Qwen2.5), with a "missed-in-Urdu" rate, content flagged as harmful in English translation but passed as normal in the original script, ranging from 2.4% to 9.9% (median 4.3%). A complete enumeration of all 205 papers across nine ALW/WOAH editions via the ACL Anthology API confirms zero dedicated Urdu papers across the entire period. Results indicate that current LLMs provide uneven safety assurance across Urdu's script varieties, with smaller open-weight models showing substantially higher instability and missed-harm rates than frontier closed models.

## 1 Introduction

The past decade of online harms research has driven the creation of benchmark datasets and evaluation frameworks through workshops such as WOAH[1] and shared tasks including HASOC,[2] SemEval,[3] and HatEval.[4] Yet this body of work contains a persistent blind spot: the near-total absence of Urdu, an Indo-Aryan language with 246 million speakers (Eberhard et al., 2023) spoken across Pakistan, India, and large diaspora communities in the United Kingdom, United States, and Gulf. Urdu speakers are highly active on social media and exposed to significant volumes of hate speech and threatening content (Akram et al., 2023; Bytes for All, Pakistan, 2014; Shafiq, 2021; Rao, 2020), yet Urdu remains largely absent from mainstream LLM safety evaluation and WOAH proceedings. WOAH 6 (2022) explicitly listed Urdu as a language urgently requiring technological support (Mathias et al., 2022); a complete enumeration of all 205 papers across all nineALW/WOAHeditionsviatheACLAnthology Python API found no work engaging substantively withUrdudata(Table2). Thispaperaddressesfour research questions:

**RQ1**: Is there sufficient work in the WOAH literature addressing online abuse and hate in Urdu over the past decade?

**RQ2**: How consistently do state-of-the-art LLMs classify hate speech when the same content is presented in Nastaliq Urdu, Roman Urdu, English translation, and code-switched Urdu-English?

**RQ3**: What dimensions of harm are systematically missing from existing Urdu hate speech datasets?

**RQ4**: Are observed differences in label instability and Missed-in-Urdu rates between models statistically significant, and do they reflect systematic label shift rather than random variation?

[1] https://www.workshoponlineabuse.com/
[2] https://hasocfire.github.io/hasoc/2021/index.html
[3] https://semeval.github.io/
[4] https://competitions.codalab.org/competitions/19935

**Contributions**: This paper makes four contributions: (i) a complete audit of ALW/WOAH proceedings (2017–2025) quantifying Urdu's representation relative to other major world languages; (ii) an empirical evaluation of five LLMs across six datasets and multiple script conditions, mea-
suring label instability and a "Missed-in-Urdu" rate; (iii) a robustness check using two independent translation sources, confirming that the observed cross-script instability is not primarily attributable to the self-translation design used in the main experiment; (iv) identification of a structural resource gap, namely the absence of a dedicated Nastaliq Urdu-English code-switched hate speech dataset; and (v) statistical validation of the observed model differences via McNemar, chisquare, and Stuart-Maxwell tests.

## 2 Background and Related Work

### 2.1 The Workshop on Online Abuse and Harms

The Workshop on Online Abuse and Harms (WOAH) traces its origins to the First Workshop on Abusive Language Online (ALW), held at ACL 2017 in Vancouver and organised by Zeerak Waseem, Wendy Hui Kyong Chung, Dirk Hovy, and Joel Tetreault (Waseem et al., 2017). The second edition, ALW2, was held at EMNLP 2018 in Brussels, with an expanded organising team that included Darja Fišer, Ruihong Huang, Vinodkumar Prabhakaran, Rob Voigt, Zeerak Waseem, and Jacqueline Wernimont (Fišer et al., 2018). By its fourth edition in 2020, the workshop had been renamed the Workshop on Online Abuse and Harms (WOAH) and was held online, organised by Seyi Akiwowo, Bertie Vidgen, Vinodkumar Prabhakaran, and Zeerak Waseem (Akiwowo et al., 2020). From this edition onward, WOAH explicitly invited submissions from civil society, particularly individuals and organisations working with women and marginalised communities disproportionately affected by online abuse, broadening the workshop's scope beyond NLP into law, psychology, sociology, and cultural studies. Now in its tenth edition, co-located with EMNLP 2026, WOAH continues to be a primary venue for computational research on detecting, classifying, and moderating abusive and harmful online content.

### 2.2 Why Urdu Matters for Online Safety

A model that successfully identifies abusive content in English while failing to detect semantically equivalent abuse in Urdu overstates its safety performance for the tenth highest used language in the world. The challenge is further complicated by Urdu's script diversity: content may appear in Nastaliq,[1] in Roman Urdu, in code-switched forms involvingEnglish, orincombinationsthereof, each of which may elicit different model behaviour. Urdu's richness does indeed challenge standard multilingual NLP pipelines, and its Nastaliq variants and usage within mainstream social media platforms are underrepresented in large-scale pretraining corpora relative to its speaker population (Nozza, 2021). Pakistan, Urdu's primary national context, has documented levels of online hate speech targeting religious minorities, political opponents, and journalists, with social media platforms identified as primary vectors (Bytes for All, Pakistan, 2014; Rao, 2020; Shafiq, 2021). The combination of a large, digitally active speaker population, a high documented incidence of online harm, and near-total absence from automated detection research makes Urdu a particularly consequential blind spot.

### 2.3 LLM-Based Hate Speech Detection

Large language models have been increasingly applied to hate speech detection in zero-shot and fewshot settings. Ghorbanpour et al. (2025) evaluated LLM prompting for hate speech detection across eight non-English languages, finding that prompt design critically affects performance and that different languages benefit from different prompting strategies. Notably, their evaluation covers five of the world's top-ten most spoken languages (Hindi, Spanish, French, Arabic, and Portuguese), yet leaves four unevaluated: Mandarin, Bengali, Russian, and Urdu. Urdu is thus absent from both the most comprehensive recent cross-lingual LLM prompting study and from the broader WOAH literature documented in Section 4.1. Melis et al. (2025) showed that prompt definition wording significantly affects zero-shot classification outcomes. Chan et al. (2024) directly suggest that translation is ineffective for code-mixed content, with detection performance degrading visibly, contextualising why evaluating models in their original Urdu script is essential rather than relying on English translation alone. Nozza (2021) showed that zeroshot cross-lingual transfer fails for low-resource languages, a finding that motivates the cross-script evaluation design of this paper. The one notable exception in Urdu is (Ahmad et al., 2025), who used an LLM for Urdu hate speech detection and outperformed BERT on both explicit and implicit hate.

[1] Nastaliq is the predominant calligraphic style of the Perso-Arabic script used to write Urdu, Persian, and other Central and South Asian languages; it is right-to-left and distinct from Roman Urdu, a Latin-script transliteration used widely on social media.

However, their work evaluates a single model on a single dataset and does not investigate cross-script consistency or the effect of script variety on classification reliability. Dey et al. (2024) examined prompting in three low-resource South Asian languages and found that translating inputs to English before classification outperformed prompting in the original language, a result that directly reflects our comparison of C1 versus C2. Vargas et al. (2024) highlight parallel challenges for Hausa, a low-resource African language, providing the closest structural analog to this work within WOAH proceedings and demonstrating that low-resource language gaps in hate speech research are a systemic rather than isolated problem.

## 3 Methodology

### 3.1 Datasets

Six datasets spanning three scripts were evaluated, summarised in Table 1. No dedicated Nastaliq Urdu-English code-switched hate speech dataset exists; the Roman Urdu-English RU-EN Emotion corpus is therefore used as the closest available proxy, and this absence is treated as a finding in its own right (Ousidhoum et al., 2019). All datasets are mapped to a unified three-class taxonomy: *Hate*, *Offensive*, and *Normal*, following prior work (Waseem and Hovy, 2016; Davidson et al., 2017). This requires converting original binary or finegrained label spaces into a consistent evaluation schema. This mapping broadly follows Offensive Language Identification Dataset (OLID)'s framing of threats and profanity as offensive-category content (Zampieri et al., 2019), with THREAT elevated to Hate on the grounds that explicit threats represent the most severe harm category in our three-way schema. We acknowledge that this is a simplification: following Zampieri et al.'s targetbased framework, the same lexical category (e.g. PROFANE) may constitute hate or mere offense depending on whether it targets a protected group, a distinction the source label spaces do not capture. Manual and automated validation of this mapping is left for future work. Each instance is evaluated under four experimental conditions designed to isolate the effects of script, transliteration, and codeswitching while holding semantic content constant. All models are applied identically across all conditions.

**(C1) Original Nastaliq Urdu.** The raw Urdu text is used in its original Nastaliq script without modification. Models classify the input directly inscript.

**(C2) English translation.** The C1 text is translated into English and classified using the same model that generated the translation. This design isolates each model's own translationclassification consistency; a robustness check using two independent translation sources is reported in (Table 3, Figure 2) and indicates this holds approximately, though not perfectly, depending on translation quality.

**(C3) Roman Urdu transliteration.** The HS-RU20 dataset provides Roman Urdu equivalents of the original Nastaliq text. This condition isolates the effect of transliteration into Latin script without translation into English.

**(C4) Code-switched Roman Urdu–English.** The RU-EN Emotion corpus is used as a proxy for Urdu–English code-switching due to the absence of a dedicated Nastaliq code-switched dataset. This condition evaluates model robustness to mixed-script inputs.

**HateXplain** is used as an English-only control dataset. Since C1 and C2 differ only in script and translation, any instability observed on HateXplain across these conditions indicates a pipeline inconsistency rather than model behaviour. The **HateInsights and Cyberbullying**

Table 1: Datasets used in the experiment. Source labels normalised to *H*/*O*/*N* where *H*=Hate, *O*=Offensive, *N*=Normal.

| Dataset Name Cited | Dataset (N=5396) | Script (Condition (C)) | Target Label | Transformation |
|---|---|---|---|---|
| HateInsights (Arshad and Shahzad, 2024) | HateInsights (N=1000) | Nastaliq (C1) | *H*/*O*/*N* | Already compatible |
| Cyberbullying (Adeeba et al., 2024) | Cyberbullying (N=916) | Nastaliq (C1) | *H*/*O*/*N* | 7-class† → *H*/*O*/*N* |
| Abusive Tweets (Amjad et al., 2022) | Abusive Tweets (N=656) | Nastaliq (C1) | *H*/*N* | Binary (1/0) → H/N |
| HS-RU-20 (Khan et al., 2022) | HS-RU-20 (N=835) | Roman Urdu (C3) | *H*/*N* | Already binary |
| RU-EN Emotion (Ilyas et al., 2023) | RU-EN Emotion (N=989) | Code-sw. (C4) | *H*/*O*/*N* | 7-class† → *H*/*O*/*N* |
| HateXplain (Mathew et al., 2021) | HateXplain (1000) | English (Control) | *H*/*O*/*N* | Already compatible |

N = Instances.† Mapping: THREAT→*H*; INSULT, OFFENSIVE, NAMECALLING, PROFANE, CURSE→*O*; NONE→ *N*.

datasets were verified to have identical label spaces but no text overlap via automated inspection and are treated as independent datasets with separate citations. As shown in Figure 6, the datasets span 2021–2024, while all evaluated models were released between mid2024 and early 2025, ensuring evaluation reflects contemporary model behaviour.

### 3.2 Models

Five LLMs are evaluated in zero-shot classification mode: GPT-4o[2], Claude Sonnet 4.5[3], Gemini 2.5 Flash[4] (Gemini Team et al., 2023), Qwen-2.5-7B-Instruct[5], and Llama-3.1-8BInstant[6] (Dubey et al., 2024). All models receive the same standardised prompt instructing classification as *Hate*, *Offensive*, or *Normal* with no in-context examples. Perspective API was considered but excluded, as it does not support Urdu, despite covering 18 other languages, a finding discussed further in the Limitations. Full prompt templates are provided in Appendix A.

### 3.3 Tools and Environment Setup

All experiments were run from a Windows machine using a dedicated Anaconda virtual environment (Python) with API access to all five model providers (OpenAI, Anthropic, Google, Together.ai, and Groq). The machine was equipped with an NVIDIA GeForce RTX 5090 GPU; however, as all five models are accessed via cloud API rather than run locally, classification was not GPUbound, and local compute was not a determining factor in experiment runtime. The principal constraint was API rate limits, particularly for the freetier Groq endpoint used for Llama-3.1. Results were written incrementally to disk with checkpointing, allowing the run to resume after interruption without data loss or duplicate API calls. [7]

[2] https://openai.com/index/ gpt-4o-system-card/

[3] https://www.anthropic.com/ claude-sonnet-4-5-system-card

[4] https://deepmind.google/technologies/gemini/flash/

[5] https://qwenlm.github.io/blog/qwen2.5/

[6] https://ai.meta.com/blog/meta-llama-3-1/

### 3.4 Experimental Conditions and Metrics

Two metrics are then computed from divergences between C1 and C2. *Label instability* is the proportion of instances that receive a different classification under C1 versus C2. *Missed-in-Urdu* is the proportion of instances classified as *Hate* or *Offensive* under C2 but as *Normal* under C1, representing content the model recognises as harmful in English but silently passes in the original Urdu script. The pipeline is illustrated in Figure 1. Full exclusion and cleaning results are provided in Appendix B within Tables 6, 7, and 8.This metric sits within a broader line of work on cross-lingual consistency and translation-based evaluation of NLP systems (Dey et al., 2024; Chan et al., 2024), which similarly finds that classification outcomes can shift substantially depending on whether inputs are evaluated in their original language or via translation.

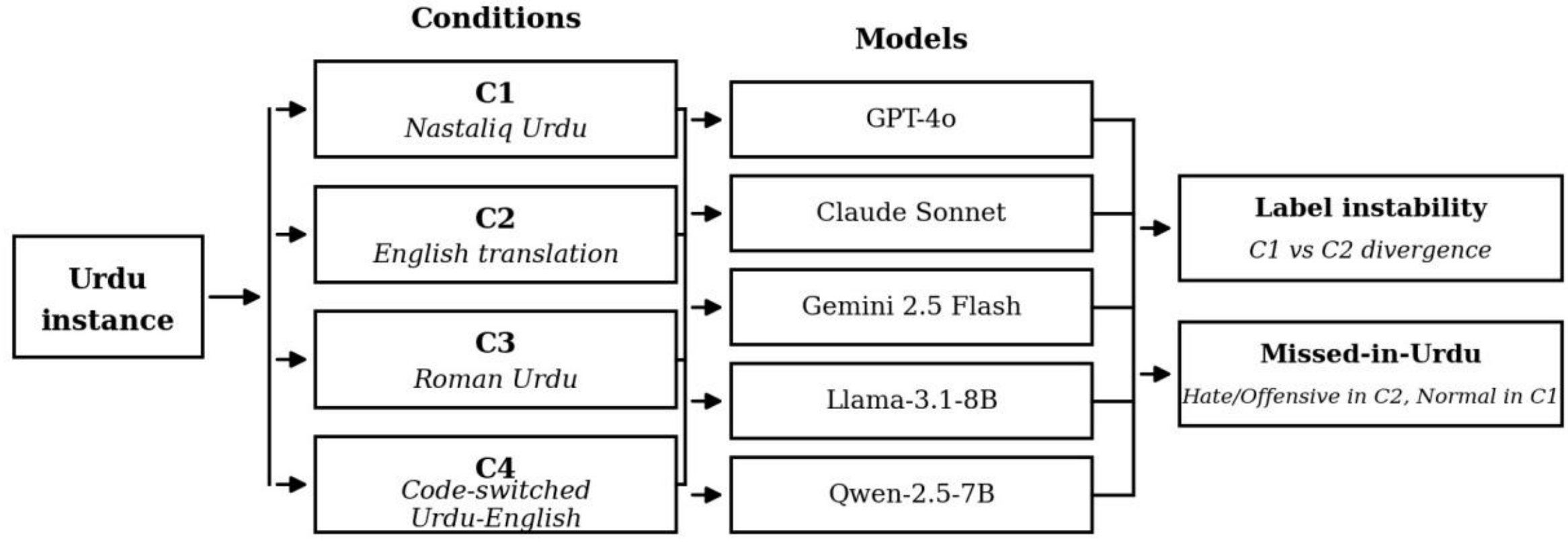

Figure 1: Four-condition classification pipeline. Each instance is classified as *Hate*, *Offensive*, or *Normal* under conditions C1–C4 per LLM sentiment classifier. Divergences between C1 and C2 yield the two evaluation metrics.

### 3.5 Statistical Significance Testing

Formal significance testing is required to confirm that observed differences between models reflect systematic behavioural variation rather than sampling artefacts. As the same text is classified by every model, observations are paired and standard independent-samples tests are inappropriate (Dror et al., 2018). Paired binary outcomes are tested by McNemar's test (McNemar, 1947): unlike a chi-square test, it conditions on discordant pairs and is sensitive to directional asymmetries in error patterns (Dietterich, 1998), and has been widely adopted for comparing NLP classifiers on shared test sets (Dror et al., 2018; Søgaard et al., 2014). For non-symmetric binary comparisons, a chi-square test on the 2×2 contingency table is appropriate (Agresti, 2002). For three-class outcomes, the Stuart-Maxwell test extends McNemar's test to square tables of arbitrary size, testing whether marginal label distributions differ significantly between paired conditions (Stuart, 1955; Maxwell, 1970). Where multiple pairwise comparisons are made simultaneously, Bonferroni correction controls the familywise error rate by dividing the significance threshold by the number of tests (Bonferroni, 1936; Dror et al., 2018).

Table 2: WOAH/ALW research attention by language: complete enumeration of all 205 papers across 9 editions via the ACL Anthology Python API, keywordmatched on titles and abstracts.

| Rank | Language | Papers |
|---|---|---|
| 1 | English | 100+ |
| 2 | Hindi (incl. Hinglish) | 6–8 |
| 3 | German | 5–6 |
| 4 | Arabic (incl. Levantine) | 4–5 |

[7] Code available at `https://anonymous.4open.science/r/WOAHJul26-047C` for review; a de-anonymised link will replace this upon acceptance.

| 5 | Italian | 4–5 |
|---|---|---|
| 6 | Dutch | 3–4 |
| 7 | Spanish (incl. Chilean) | 2–3 |
| 8 | Portuguese | 2–3 |
| 9 | Mandarin / French | 1–2 |
| 10 | Russian | 1–2 |
| – | **Bengali** | **0** |
| – | **Urdu** | **0** |

# 4 Results

Results are organised by research question: Section 4.1 addresses WOAH coverage of Urdu (RQ1), Sections 4.2 -I and 4.3 -II address cross-script classification consistency across all four experimental conditions as well as extra robustness checks, Section 4.4 missing harm dimensions in existing Urdu resources (RQ3), and Section 4.5 statistical significance of the observed model differences (RQ4).

## 4.1 Coverage of Urdu in WOAH Literature (RQ1)

A complete enumeration of all 205 papers across nine ALW/WOAH editions (2017–2025) was conducted via the ACL Anthology Python API,[8] retrieving titles and abstracts across all nine volume IDs and searching for language mentions using a controlled keyword set. This constitutes a complete enumeration of the indexed record rather than a practitioner-style search; so a paper using Urdu data without mentioning it in the abstract would not be surfaced, though such a paper would equally not be discoverable by practitioners searching the literature. The results confirm that Urdu has appeared in no dedicated paper across the entire period reviewed, and Bengali, the world's seventh most spoken language, is equally absent (Table 2). Several languages with far fewer speakers, including German (rank 17), Italian (rank 22), and Dutch (rank 32), each have multiple dedicated WOAHpapers, reflectingthefield'shistoricalroots in European-language NLP communities. Notably, WOAH 6 (2022) explicitly named Urdu alongside Yoruba and Amharic as a language urgently requiring technological support (Mathias et al., 2022); despite this invitation, no dedicated Urdu paper appeared in that or any subsequent edition. There is, in short, no sufficient body of work in the WOAH literature addressing Urdu online abuse across the decade reviewed.

> *Takeaway.* Urdu, spoken by 246 million people, has received zero dedicated papers across nine ALW/WOAH editions despite an explicit invitation in 2022 reflecting a gap of uneven research attention and geographical reach.

## 4.2 Cross-Script Classification Consistency (RQ2 - I)

Table 4 presents instability and Missed-in-Urdu rates across all five models ($N$ =4,531 to 4,553 per model). HateXplain confirms 0.0% on both metrics by construction, ruling out pipeline errors. See Figure 8 in Appendix B. Between 15.9% and 31.6% of instances receive a different label depending purely on script. Frontier models cluster between 15.9% and 18.0%; Llama-3.1 and Qwen-2.5 reach 27.3% and 31.6%, roughly double the frontier rate (Figure 3). Figure 4 also summarises this per-model split directly against the cross-model medians. Every model also misses harmful Urdu content it correctly flags in English translation: Missed-in-Urdu rates range from 2.4% (GPT-4o) to 9.9% (Qwen-2.5). This is not a failure of comprehension but of script-conditioned safety behaviour. Figure 7 (Appendix B) shows that open-weight models produce visibly wider crossflows from *Hate* and *Offensive* in C1 into *Normal* in C2. The statistical reliability of these differences is confirmed in Figure 5 and discussed further in Section 4.5. To further test whether the selftranslation design (Section 3.1) affects these instability estimates, a stratified subset of 125 instances was re-evaluatedusing two independent translation sources: NLLB-200-distilled-600M (NLLB Team et al., 2022) and Google Cloud Translate, on the three frontier models still accessible via API at the time of writing. Qwen-2.5-7B and Llama-3.18B could not be re-tested, as their Groq and Together.ai endpoints were deprecated after the original experiment was run (see Limitations). Table 3 and Figure 2 show that the two independent sources diverge from one another as much as they diverge from self-translation. Google Translate's instability rates (12.0–15.2%) are close to the original self-translation rates (9.6–14.4%), while NLLB-200 produces substantially higher instability (19.2–24.8%) for every model tested. NLLB and Google agree with each other on the resulting classification label only 76.8–86.4% of the time across models, indicating that translation *system quality*, rather than whether a model translates its own input, is a material source of variation in these estimates. Because a high-quality commercial MT system (Google Translate) produces instability comparable to self-translation, we do not find evidence that the original self-translation design substantially inflated the paper's central instability estimates; the higher NLLB figures likely reflect that system's lower translation quality on Urdu rather than a translation-source confound specific to self-translation. We nonetheless flag translation system choice as a factor that should be controlled for and reported in future cross-script safety evaluations.

[8] https://pypi.org/project/acl-anthology/

| Model | N | Instability (%) | | |
|---|---|---|---|---|
| | | Self | NLLB | Google |
| GPT-4o | 125 | 14.4 | 24.8 | 12.0 |
| Claude Sonnet 4.5 | 125 | 9.6 | 19.2 | 12.0 |
| Gemini 2.5 Flash | 125 | 12.8 | 24.0 | 15.2 |
| | | Missed-in-Urdu (%) | | |
| | | Self | NLLB | Google |
| GPT-4o | 125 | 3.2 | 5.6 | 2.4 |
| Claude Sonnet 4.5 | 125 | 3.2 | 1.6 | 1.6 |
| Gemini 2.5 Flash | 125 | 4.8 | 4.0 | 3.2 |

Table 3: Label instability and Missed-in-Urdu rates under self-translation (original design) versus two independent translation sources (NLLB-200 and Google Translate), on a stratified 125-instance subset. NLLB– Google label agreement ranged from 76.8% (Gemini) to 86.4% (Claude), indicating the two independent sources do not fully agree with each other either.

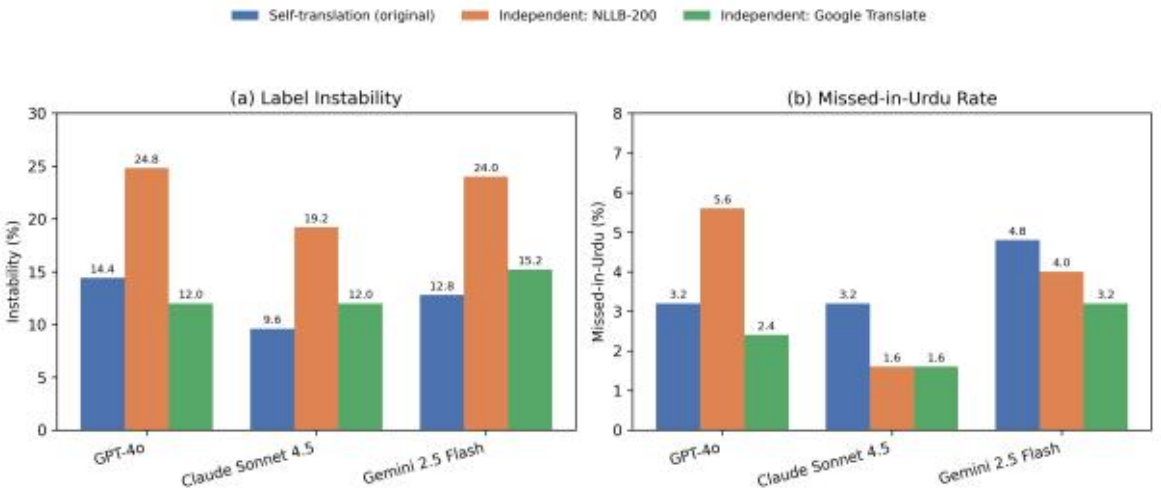

Figure 2: Comparison of label instability (a) and Missed-in-Urdu rate (b) under self-translation, NLLB200, and Google Translate, per model, on the 125instance robustness subset.

*Takeaway.* Every model tested assigns different moderation labels to the same content depending on whether it is presented in Nastaliq Urdu or English translation, with median instability of 18.0% (SD=7.1%) and median "Missed-inUrdu" of 4.3% (SD=2.9%) across all models and datasets; a robustness check using two independent translation sources indicates these estimates are not an artefact of self-translation, though the magnitude is sensitive to translation quality.

### 4.3 Transliteration and Code-Switching Effects (RQ2 - II)

RQ2 also asked how models handle Roman Urdu (C3) and code-switched input (C4); this section addresses that comparison directly. Beyond the absence of a dedicated resource, the available proxy data offers a partial signal on how transliteration and code-switching affect model consistency. HS-RU-20 (C3, Roman Urdu transliteration) and RU-EN Emotion (C4, Code-switched Roman Urdu–English) show the lowest instability rates of any dataset for every model tested (See Figure 2), including the frontier models, suggesting that Latin-script transliteration, whether monolingual or mixed with English, does seem to narrow the cross-script gap relative to Nastaliq. This may suggest that the smaller gap reflects greater lexical and tokenisation overlap between Roman Urdu and English than between Nastaliq Urdu and English, though the sample is limited to a single dataset per conditionandRU-ENEmotionusesemotionlabels rather than harm labels, so this pattern should be treated as suggestive rather than conclusive, and needs to be explored further.

*Takeaway.* Available proxy data also suggests Latin-script transliteration and code-switching are associated with lower cross-script instability than Nastaliq script, though this pattern rests on a singledatasetperconditionandshouldbetreated as suggestive rather than conclusive, needing more study.

### 4.4 Missing Harm Dimensions in Urdu Resources (RQ3)

The experimental design itself surfaces a structural absence: nodedicatedNastaliqUrdu-

Englishcodeswitched hate speech dataset exists. The RU-EN Emotion corpus is the closest available proxy, using Roman Urdu rather than Nastaliq and emotion rather than harm labels. Code-switching between Urdu and English is pervasive in Pakistani social media; people naturally mix English and
Urdu words when discussing digital or modern concepts. For example: “çŬó˒Ěōó ÿŎ ì àṇó< meeting attend ki¿‘ (Did you attend today’s meeting?) Or should we reschedule?”. Content moderation systems that process only monolingual input are structurally unable to evaluate it. The absence of a dedicated resource means that the field cannot currently measure how well any model handles this variety, a limitation that affects every existing evaluation in this space, not only this paper.

A related concern is annotation reliability. Where datasets are annotated by small or homogeneous groups, shared cultural assumptions can propagate into labels. Across the full run, 1,216 instances were labelled *Hate* by the original annotators but *Normal* by all five models under C1. A single pilot-stage example of a content defending a religious minority annotated as *Hate* raises the possibility that some cases reflect annotation bias rather than model failure, though this is illustrative rather than a validated pattern. Whether this holds at scale requires qualitative review and is left for future work; without it, and possible annotation bias in source corpora propagate into downstream evaluation.

Table 4: Label instability [C1 vs. C2 (%)] by dataset and model. HateXplain (English gold standard) shows 0.0% across all five models by construction, confirming that instability on the remaining five datasets is attributable to language and script rather than any pipeline issue. Instability values are shown to one decimal place.

| **Model** | **Instability** | **Missed-in-Urdu** |
|---|---|---|
| GPT-4o | 18.0% | 2.4% |
| Claude Sonnet 4.5 | 16.3% | 3.6% |
| Gemini 2.5 Flash | 15.9% | 4.3% |
| Qwen-2.5-7B | 31.6% | 9.9% |
| Llama-3.1-8B | 27.3% | 6.5% |
| *Median (SD)* | *18.0 (7.1)%* | *4.3 (2.9)%* |

*Takeaway.* The absence of a dedicated Nastaliq Urdu-English code-switched hate speech dataset is itself a measurable gap, and the pattern of model-annotation disagreement suggests existing gold annotations may warrant qualitative review for possible annotation bias before use as ground truth and needs more proactive research.

### 4.5 Statistical Significance Findings (RQ4)

Three significance tests were applied to confirm that observed differences are systematic rather than artefacts of sampling. Pairwise instability was assessed using McNemar’s test (paired binary outcomes); pairwise Missed-in-Urdu using chi-square tests; and C1-vs-C2 label shift per model using the Stuart-Maxwell test (the three-class generalisation of McNemar’s). All pairwise tests are Bonferronicorrected ($\alpha = 0.005$, ten comparisons); results are reported in Appendix B within Tables 9, 10, and 5. The three frontier models form a statistically indistinguishable cluster on instability; every frontier-vs-open-weight comparison is highly significant ($p < 0.0005$), confirming the twotier structure in Table 4. GPT-4o shows significantly lower Missed-in-Urdu than Claude Sonnet 4.5 ($p < 0.0005$) despite similar instability, suggesting greater conservatism about passing harmful Urdu content. Stuart-Maxwell tests confirm that label distributions shift significantly between
C1 and C2 for every model tested ($p < 0.01$ or stronger).

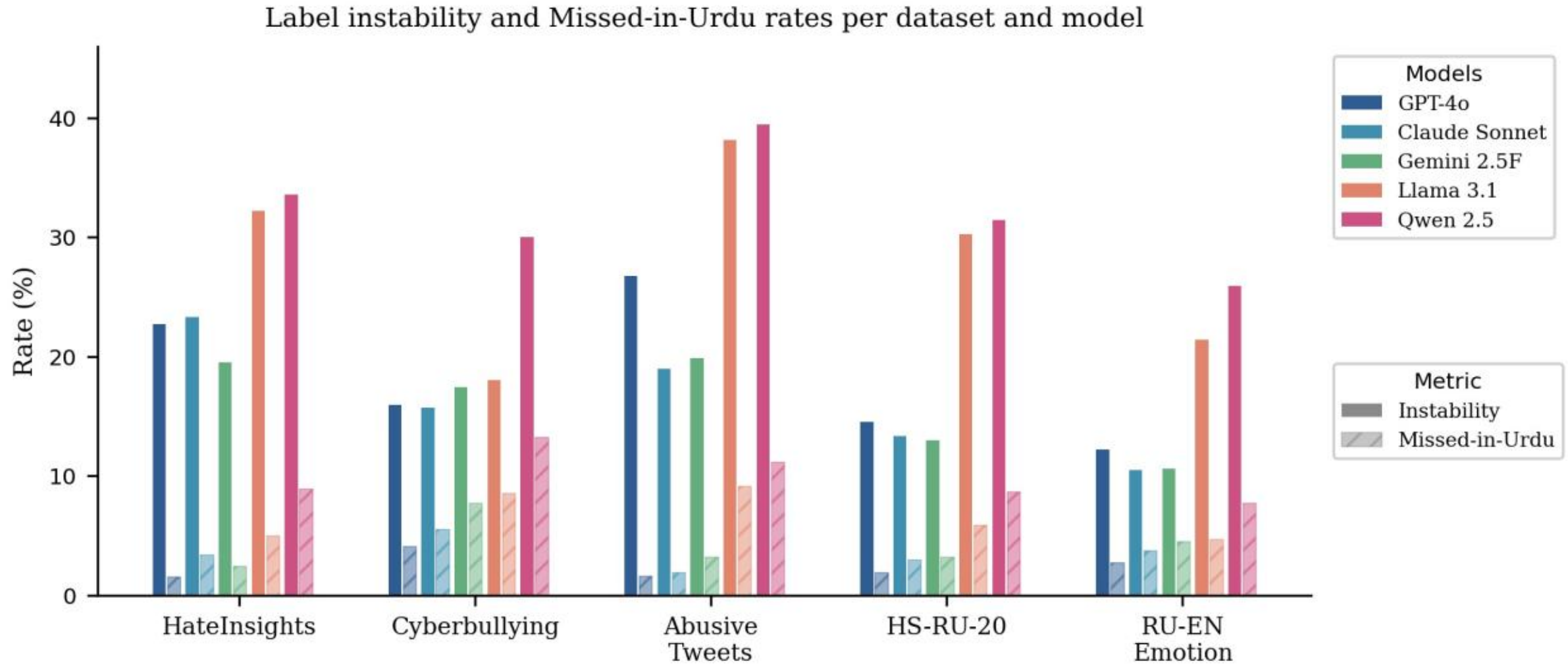

Figure 3: Label instability and Missed-in-Urdu rates per dataset and model. Each dataset group shows paired bars per model: solid (instability) and hatched (Missed-in-Urdu). Frontier models (GPT-4o, Claude Sonnet 4.5, Gemini 2.5 Flash) shown in blue–green; open-weight models (Llama-3.1, Qwen-2.5) in warm tones.

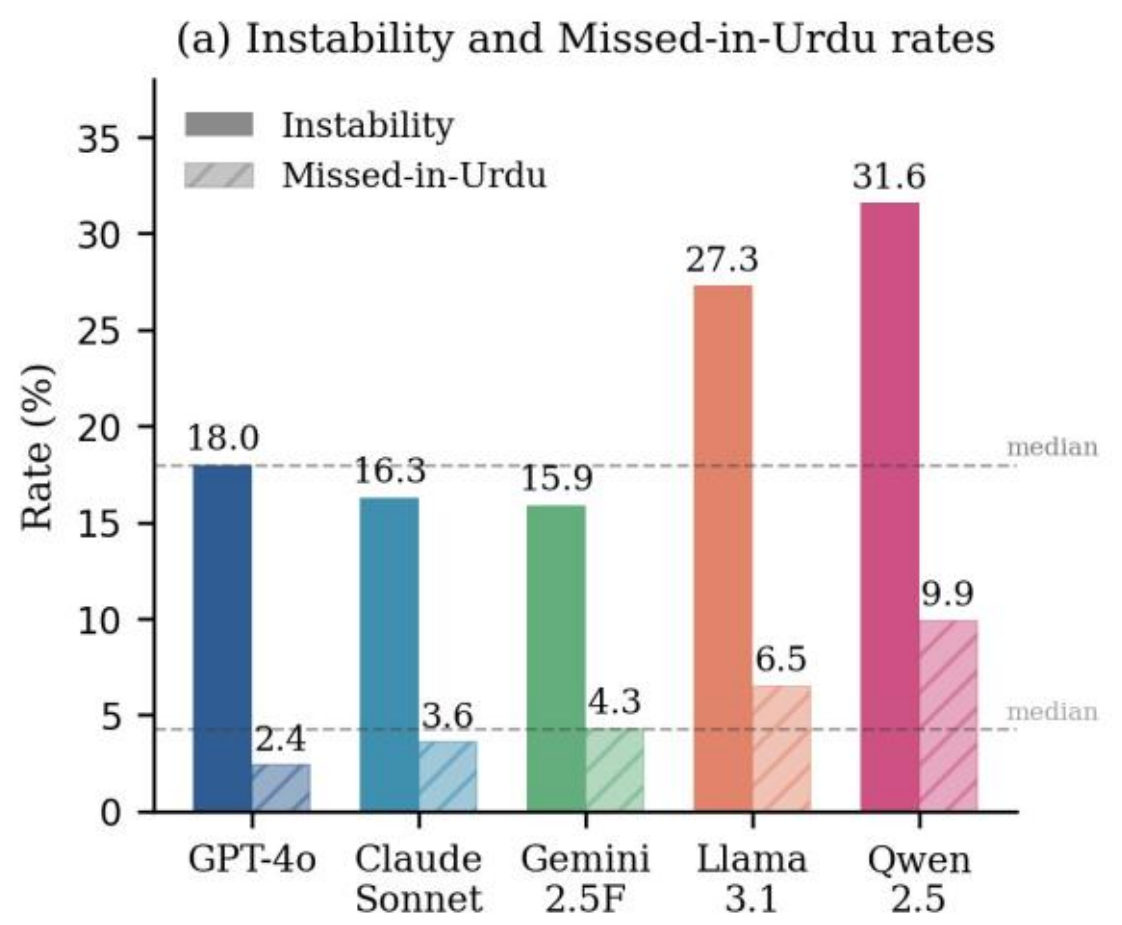

Figure 4: Label instability and Missed-in-Urdu rates per model after exclusion of error and refusal rows ($N$ =4,531–4,553). Dashedlinesshowcross-modelmedians (18.0% and 4.3%).

*Takeaway.* These statistical tests rule out the possibility that the observed instability is attributable to random label noise; the results in Table 4 can be considered reliable.

(b) Pairwise significance tests

| | GPT-4o | Claude | Gemini | Llama | Qwen |
|---|---|---|---|---|---|
| GPT-4o | | n.s. | p=0.003 | p<.0001 | p<.0001 |
| Claude | p<.0001 | | n.s. | p<.0001 | p<.0001 |
| Gemini | n.s. | n.s. | | p<.0001 | p<.0001 |
| Llama | p=0.0003 | n.s. | n.s. | | p<.0001 |
| Qwen | p=0.005 | p=0.002 | p=0.0002 | p=0.0001 | |

U — L

Not significant ($\alpha$ = 0.005, Bonferroni) • U: Instability (McNemar) • L: Missed-in-Urdu ($\chi^2$)

Figure 5: Pairwise significance matrix: upper triangle (U) shows McNemar test results for instability; lower triangle (L) shows chi-square results for Missed-in-Urdu. Grey diagonal cells are self-comparisons. White cells indicate non-significance after Bonferroni correction ($\alpha$ = 0.005, 10 comparisons). P-values shown for all significant pairs.

## 5 Conclusion

Across five Urdu-script datasets and five models, label instability ranges from 15.9% to 31.6% and Missed-in-Urdu rates from 2.4% to 9.9%, with open-weight models performing substantially worse than frontier models on both measures. A robustness check using two independent translation sources confirmed that these instability estimates are not primarily a result of the original self-translation design, though their magnitude is sensitive to translation system quality. Transliteration and code-switching showed the opposite pattern to Nastaliq-only comparisons: Roman Urdu (C3) and code-switched input (C4) produced the lowest instability of any condition across all five models, suggesting that Latin-script forms narrow rather than widen the cross-script safety gap, though this observation rests on a single dataset per condition. A complete enumeration of all 205 ALW/WOAHpapersconfirmszerodedicatedUrdu papers across nine editions, despite Urdu being the world's tenth most spoken language and despite an explicit invitation from WOAH 6 (2022). The field consequently has no Urdu hate speech benchmark, no shared evaluation protocol, and no established dataset for system comparison. Addressing this requires dedicated Nastaliq Urdu inclusiveEnglish code-switched datasets, Urdu-inclusive safety benchmarks, and a research community that treats absence of coverage as a measurable safety risk. The 246 million Urdu speakers exposed to undetected online harm are not a niche population: they are the tenth largest linguistic community in the world.

## Limitations

Sample sizes range from $N = 700$ (Abusive Tweets) to $N = 1{,}000$ per dataset; HS-RU-20 reached 85% of target due to an interrupted run and is reported at $N = 853$–854. The translation step (C2) is performed by the same model under test, isolating each model's own consistency but meaning results cannot distinguish translation quality from classification consistency. Binary source datasets (Abusive Tweets, HS-RU-20) have no *Offensive* class, which may inflate apparent instability for those datasets. The 1,216 instances where models disagreed with gold *Hate* annotations have not been manually reviewed; whether these reflect model failure, annotation bias, or both remains an open question for future work. The RU-EN Emotion corpus is used as a code-switched proxy with emotion labels mapped to harm labels, which is an approximation. The ACL Anthology audit operates on titles and abstracts only; a paper not mentioning Urdu in its abstract would not be surfaced, though it would equally not be discoverable by practitioners. All models are accessed via commercial API; reproducibility depends on provider versioning and is not guaranteed over time. Perspective API does not support Urdu despite covering 18 other languages, itself part of the motivation for this paper. This risk materialised during revision: Groq deprecated `llama-3.1-8b-instant` in August 2026, and Together.ai's serverless endpoint for `Qwen2.5-7B-Instruct-Turbo` became inaccessible under the free tier, both after the original experiment was conducted. As a result, the independent-translation robustness check in Section 4.2 (Table 3, Figure 2) could only be re-run on the three frontier models; the original five-model results in Table 4 remain unaffected.

## Future Work

Future work should validate the THREAT/PROFANE-to-label mapping (Section 3.1) through target-aware relabelling, either manual review or an LLM-based check of whether flagged content specifically targets a protected group, following Zampieri et al. (2019)'s targetbased framework.

## Ethical Considerations and use of AI

Alldatasetsarepubliclyavailableoravailableupon registration; no personally identifying information is stored or reported. Abusive and hateful content is analysed for research purposes only, following established community guidelines. Claude Sonnet4.6(Anthropic)assistedwithLaTeXformatti ng and figure generation; all research design, analysis, interpretation, and conclusions are the authors' own.

## Acknowledgements

The authors would like to thank Dr Vaibhav Bajpai of the Hasso Plattner Institute for his valuable suggestions, insightful feedback, and careful review, which helped strengthen this paper. The authors also thank Dr Asegul Hulus for helpful initial discussions and encouragement.

## References

Farah Adeeba, Muhammad Irfan Yousuf, Izza Anwer, Sardar Umair Tariq, Abdullah Ashfaq, and Malik Naqeeb. 2024. Addressing cyberbullying in Urdu tweets: a comprehensive dataset and detection system. *PeerJ Computer Science*, 10:e1963.

Alan Agresti. 2002. *Categorical Data Analysis*, 2nd edition. John Wiley & Sons.

Muhammad Ahmad, Muhammad Usman, Sulaiman Khan, Muhammad Muzamil, Ameer Hamza, Muhammad Jalal, Ildar Batyrshin, Usman Sardar, and Carlos Aguilar-Ibañez. 2025. Hate speech detection using social media discourse: A multilingual approach with large language model. *African Journal of Biomedical Research*, 28(2S):321–328.

Seyi Akiwowo, Bertie Vidgen, Vinodkumar Prabhakaran, and Zeerak Waseem, editors. 2020. *Proceedings of the Fourth Workshop on Online Abuse and Harms*. Association for Computational Linguistics, Online.

Muhammad Hammad Akram, Khurram Shahzad, and Maryam Bashir. 2023. ISE-Hate: A benchmark corpus for inter-faith, sectarian, and ethnic hatred detection on social media in Urdu. *Information Processing & Management*, 60(3):103270.

Maaz Amjad, Noman Ashraf, Grigori Sidorov, Alisa Zhila, Liliana Chanona-Hernandez, and Alexander Gelbukh. 2022. Automatic abusive language detection in Urdu tweets. *Acta Polytechnica Hungarica*, 19(10):143.

Muhammad Umair Arshad and Waseem Shahzad. 2024. Understanding hate speech: the HateInsights dataset and model interpretability. *PeerJ Computer Science*, 10:e2372.

Carlo Emilio Bonferroni. 1936. Teoria statistica delle classi e calcolo delle probabilità. *Pubblicazioni del R. Istituto Superiore di Scienze Economiche e Commerciali di Firenze*, 8:3–62.

Bytes for All, Pakistan. 2014. Hate speech: A study of Pakistan's cyberspace. Technical report, Association for Progressive Communications.

Fai Leui Chan, Duke Nguyen, and Aditya Joshi. 2024. "Is Hate Lost in Translation?": Evaluation of multilingual LGBTQIA+ hate speech detection. *arXiv preprint arXiv:2410.11230*.

Thomas Davidson, Dana Warmsley, Michael Macy, and Ingmar Weber. 2017. Automated hate speech detection and the problem of offensive language. In *Proceedings of the International AAAI Conference on Web and Social Media*, volume 11, pages 512–515.

Krishno Dey, Prerona Tarannum, Md. Arid Hasan, Imran Razzak, and Usman Naseem. 2024. Better to ask in English: Evaluation of large language models on English, low-resource and cross-lingual settings. *arXiv preprint arXiv:2410.13153*.

Thomas G. Dietterich. 1998. Approximate statistical tests for comparing supervised classification learning algorithms. *Neural Computation*, 10(7):1895–1923.

Rotem Dror, Gili Baumer, Segev Shlomov, and Roi Reichart. 2018. The hitchhiker's guide to testing statistical significance in natural language processing. In *Proceedings of the 56th Annual Meeting of the Association for Computational Linguistics*, pages 1383–1392. Association for Computational Linguistics.

Abhimanyu Dubey, Abhinav Jauhri, Abhinav Pandey, Abhishek Kadian, Ahmad Al-Dahle, Aiesha Letman, Akhil Schelten Mathur, and et al. 2024. The Llama 3 herd of models. *Preprint*, arXiv:2407.21783.

David M. Eberhard, Gary F. Simons, and Charles D. Fennig. 2023. *Ethnologue: Languages of the World*, 26th edition. SIL International.

Darja Fišer, Ruihong Huang, Vinodkumar Prabhakaran, Rob Voigt, Zeerak Waseem, and Jacqueline Wernimont, editors. 2018. *Proceedings of the 2nd Workshop on Abusive Language Online (ALW2)*. Association for Computational Linguistics, Brussels, Belgium.

Gemini Team, Rohan Anil, Sebastian Borgeaud, Yonghui Wu, Jean-Baptiste Alayrac, Jiahui Yu, Radu Soricut, Johan Schalkwyk, Andrew M. Dai, and Anja Hauth et al. 2023. Gemini: A family of highly capable multimodal models. *Preprint*, arXiv:2312.11805.

Faeze Ghorbanpour, Daryna Dementieva, and Alexander Fraser. 2025. Can prompting LLMs unlock hate speech detection across languages? A zero-shot and few-shot study. In *Proceedings ofthe9thWorkshoponOnlineAbuseandHarms (WOAH)*, pages 413–425. Association for Computational Linguistics.

Abdullah Ilyas, Khurram Shahzad, and Muhammad Kamran Malik. 2023. Emotion detection in code-mixed Roman Urdu English text. *ACM Transactions on Asian and Low-Resource Language Information Processing*, 22(2).

Muhammad Moin Khan, Khurram Shahzad, and Muhammad Kamran Malik. 2022. Hate speech detection in Roman Urdu. *ACM Transactions on Asian and Low-Resource Language Information Processing*, 20(1):1–19.

Binny Mathew, Punyajoy Saha, Seid Muhie Yimam, Chris Biemann, Pawan Goyal, and Animesh Mukherjee. 2021. HateXplain: A benchmark dataset for explainable hate speech detection. In *Proceedings of the AAAI Conference on Artificial Intelligence*, volume 35, pages 14867– 14875.

Lambert Mathias, Bertie Vidgen, Zeerak Waseem, Aida Davani, and Kanika Narang. 2022. The 6th workshop on online abuse and harms: Call for papers. ACL Member Portal. Co-located with NAACL 2022, Seattle. Theme: On Developing Resources and Technologies for Low Resource Online Abuse and Harms.

A. E. Maxwell. 1970. Comparing the classification of subjects by two independent judges. *British Journal of Psychiatry*, 116(535):651–655.

QuinnMcNemar.1947. Noteonthesamplingerror of the difference between correlated proportions or percentages. *Psychometrika*, 12(2):153–157.

Matteo Melis, Gabriella Lapesa, and Dennis Assenmacher. 2025. A modular taxonomy for hate speech definitions and its impact on zero-shot LLM classification performance. In *Proceedings of the 9th Workshop on Online Abuse and Harms (WOAH)*. Association for Computational Linguistics.

NLLB Team, Marta R. Costa-jussà, James Cross, Onur Çelebi, Maha Elbayad, Kenneth Heafield, Kevin Heffernan, Elahe Kalbassi, Janice Lam, Daniel Licht, Jean Maillard, and et al. 2022. No language left behind: Scaling humancentered machine translation. *arXiv preprint arXiv:2207.04672*.

Debora Nozza. 2021. Exposing the limits of zeroshot cross-lingual hate speech detection. In *Proceedings of the 59th Annual Meeting of the ACL*.

Nedjma Ousidhoum, Zizheng Lin, Hongming Zhang, Yangqiu Song, and Dit-Yan Yeung. 2019. Multilingual and multi-aspect hate speech analysis. In *Proceedings of the 2019 Conference on Empirical Methods in Natural Language Processing and the 9th International Joint Conference on Natural Language Processing (EMNLPIJCNLP)*, pages 4675–4684, Hong Kong, China. Association for Computational Linguistics.

Muhammad Furqan Rao. 2020. Hate speech and media information literacy in the digital age: A case study of 2018 elections in Pakistan. *Global Media Journal*, 18(34):202.

Muhammad Osama Shafiq. 2021. Hate speeches on Twitter and Facebook in south asia: A case study of Malala Yousufzai. *Pakistan Journal of International Affairs*, 4(1).

Anders Søgaard, Anders Johannsen, Barbara Plank, Dirk Hovy, and Hector Martínez Alonso. 2014. What's in a p-value in NLP? In *Proceedings of the 18th Conference on Computational Natural Language Learning*, pages 1–10. Association for Computational Linguistics.

Alan Stuart. 1955. A test for homogeneity of the marginal distributions in a two-way classification. *Biometrika*, 42(3–4):412–416.

Francielle Vargas, Samuel Guimarães, Shamsuddeen Hassan Muhammad, Diego Alves, Ibrahim Said Ahmad, Idris Abdulmumin, Diallo Mohamed, Thiago Pardo, and Fabrício Benevenuto. 2024. HausaHate: Expert annotated corpus for Hausa hate speech. In *Proceedings of the 8th Workshop on Online Abuse and Harms (WOAH)*. Association for Computational Linguistics.

Zeerak Waseem, Wendy Hui Kyong Chung, Dirk Hovy, and Joel Tetreault, editors. 2017.

*Proceedings of the First Workshop on Abusive Language Online*. Association for Computational Linguistics, Vancouver, BC, Canada.

Zeerak Waseem and Dirk Hovy. 2016. Hateful symbols or hateful people? Predictive features for hate speech detection on Twitter. In *Proceedings of the NAACL Student Research Workshop*, pages 88–93.

Marcos Zampieri, Shervin Malmasi, Preslav Nakov, Sara Rosenthal, Noura Farra, and Ritesh Kumar. 2019. Predicting the type and target of offensive posts in social media. In *Proceedings of the 2019 Conference of the North American Chapter of the Association for Computational Linguistics: Human Language Technologies, Volume 1 (Long and Short Papers)*, pages 1415–1420.

# A Prompt Templates

All models were queried via their respective APIs. Twoprompttemplateswereusedthroughouttheexperiment, applied identically across all five models and all datasets.

**Classification prompt.** The following system prompt was used for conditions C1, C2, C3, and C4:

> *You are a content moderation assistant. Classify the social media post as one of: HATE = targets a person or group with hatred; OFFENSIVE = rude or insulting but not identity-based hate; NORMAL = neither hate nor offensive. Reply with exactly one word: HATE, OFFENSIVE, or NORMAL.*

Theuserturnwasformattedas: `Post: "{text}" \n\nLabel:`

**Translation prompt (C2 only).** The following system prompt was used to produce the English translation under condition C2, applied by the same model under test rather than an external translation service:

> *You are a professional translator. TranslatethefollowingUrdusocialmedia post to English. Preserve the tone, intensity, and meaning exactly — do not soften or sanitise. Reply with only the English translation, nothing else.*

The user turn consisted of the raw Nastaliq Urdu text. Label parsing extracted the first occurrence of HATE, OFFENSIVE, or NORMAL from the model response; responses containing none of these tokens were recorded as REFUSED and excluded from instability calculations.

# B Additional Experimental Details

This appendix presents the additional result tables and figures discussed in the main text.

Table5: Stuart–MaxwelltestresultsforC1versusC2label shift per model (three classes: *Hate*, *Offensive*, and *Normal*; $\alpha$ =0.05).

| Model | $\chi^2$ | df | $p$ |
|---|---|---|---|
| GPT-4o | | | $< .001^{***}$ |
| Llama-3.1 | | | 001 |
| Qwen-2.5 | | | $< .001^{***}$ |

$^{**}p < .01$; $^{***}p < .001$

313.78 2

$< .001^{***}$

Claude Sonnet 4.5106.69 ** 2

Gemini 2.5 Flash 9.63 2 .008

145.35 2 <. ***

22.94 2

The test determines whether the marginal distribution of labels shifts significantly between Nastaliq Urdu (C1) and its English translation (C2).

Table 6: Error and refusal rows excluded before analysis.

| Model | Error rows | Refusal rows |
|---|---|---|
| GPT-4o | 1 | 23 |
| Claude Sonnet 4.5 | 1 | 2 |
| Gemini 2.5 Flash | 10 | 2 |
| Llama-3.1-8B | 0 | 2 |
| Qwen-2.5-7B | 1 | 0 |
| *Total* | *13* | *29* |

*Error rows* indicate API failures. *Refusal rows* indicate cases in which the C2 translation returned a model refusal rather than a translation. Both types were excluded from all analyses.

Table 7: Clean instance counts after exclusion of error rows, refusal rows, and the HateXplain pipeline control dataset.

| Model | Clean rows | Excluded |
|---|---|---|
| GPT-4o | 4,531 | 33 |
| Claude Sonnet 4.5 | 4,552 | 12 |
| Gemini 2.5 Flash | 4,545 | 19 |
| Llama-3.1-8B | 4,551 | 13 |
| Qwen-2.5-7B | 4,553 | 11 |
| *Total* | *22,732* | *88* |

Table 8: Unique source texts per dataset after cleaning (model-agnostic). Totalacrossallmodels=uniquetexts × 5, subject to per-model exclusions.

| **Dataset** | **Unique texts** | **All models** |
|---|---|---|
| HateInsights | 1,000 | 4,994 |
| Cyberbullying | 916 | 4,993 |
| RU-EN Emotion | 989 | 4,996 |
| HS-RU-20 | 835 | 4,266 |
| Abusive Tweets | 656 | 3,483 |
| *Total* | *4,396* | *22,732* |

*HS-RU-20* reached 85% of the $N$ =1,000 target due to an interrupted run. *Abusive Tweets* was limited by available stratified data ($N$ =700 target).

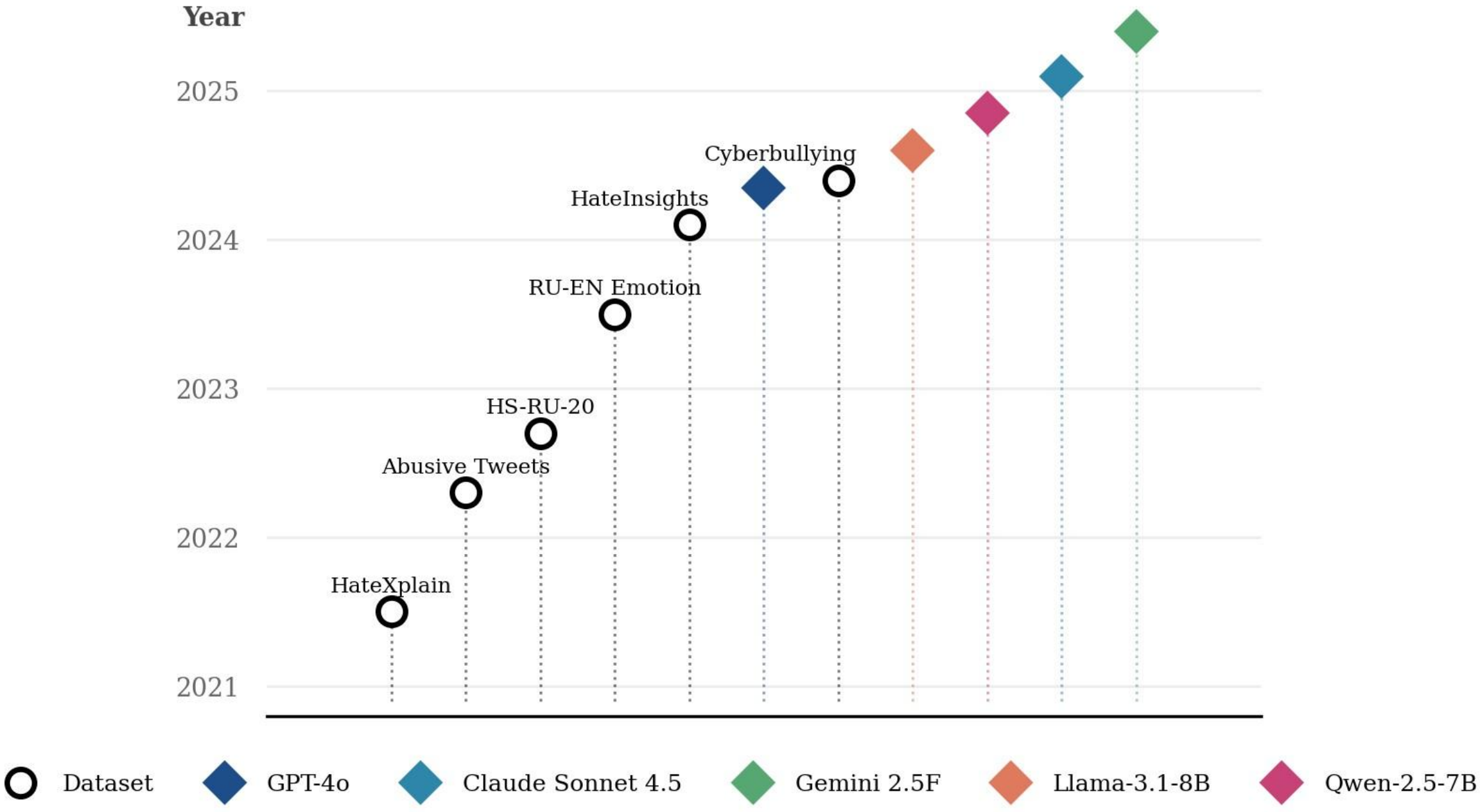

Figure 6: Release timeline of datasets (circles) and models (diamonds) used in this study. All datasets were published between 2021 and 2024, and all models between mid-2024 and early 2025, confirming that the evaluation reflects the current state of the art. Model colours match those used throughout the paper.

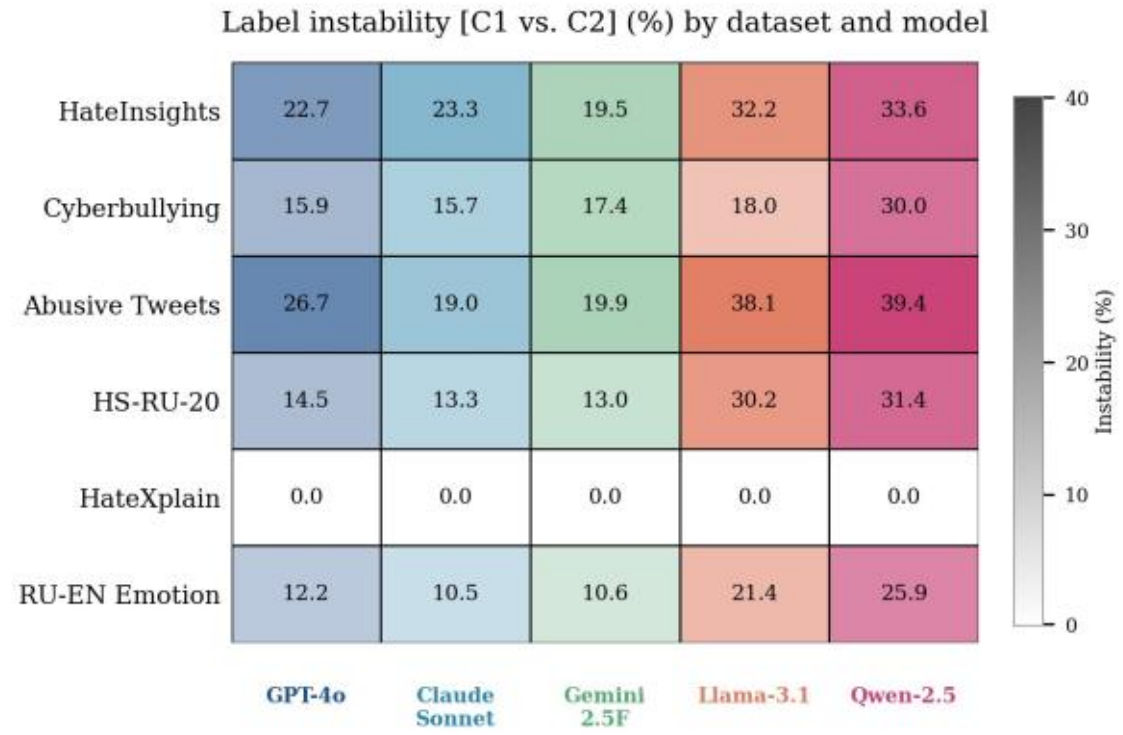

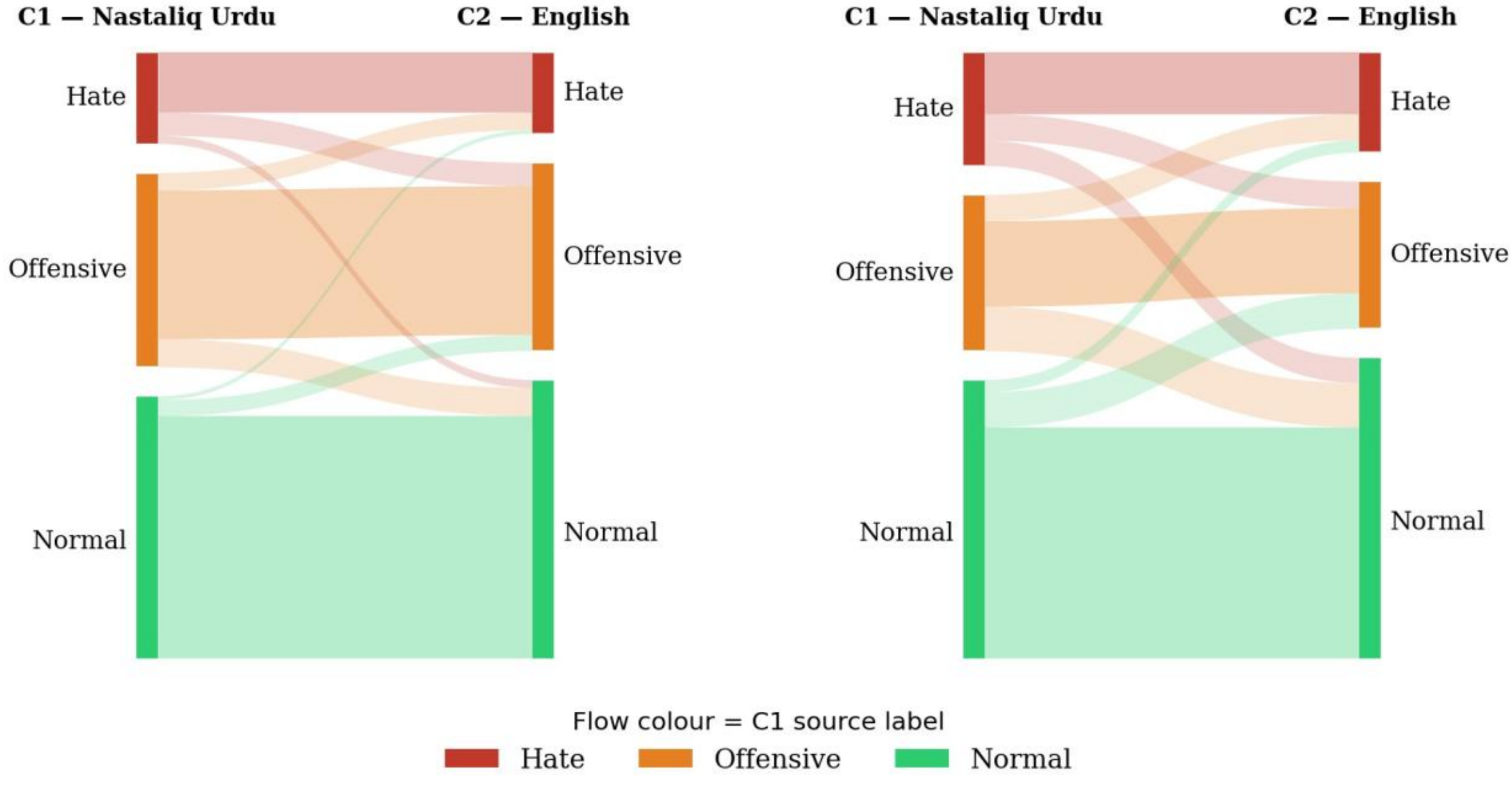

Figure 7: Label flow from C1 (Nastaliq Urdu) to C2 (English translation) for frontier models (left) and openweight models (right), aggregated across all five Urdu-script datasets. Flow colour indicates the C1 source label. Cross-flows represent label changes; wider cross-flows indicate greater instability. Open-weight models show more cross-flow, particularly from *Hate* and *Offensive* to *Normal*.

Figure 8: Label instability (C1 versus C2, %) by dataset and model. HateXplain, the English gold standard, shows 0.0% across all five models by construction, confirming that instability in the remaining five datasets is attributable to language and script.

Table 9: McNemar test results for pairwise label instability (Bonferroni-corrected, $\alpha = 0.005$, $N = 4{,}365$ matched instances).

| Model A | Model B | Stat | $p$ |
|---|---|---|---|
| Claude Sonnet 4.5 | GPT-4o | 3.82 | 0.051 |
| Claude Sonnet 4.5 | Gemini 2.5 Flash | 1.64 | 0.201 |
| GPT-4o | Gemini 2.5 Flash | 9.02 | 0.003* |
| Claude Sonnet 4.5 | Llama-3.1 | 171.22 | < .001*** |
| Claude Sonnet 4.5 | Qwen-2.5 | 286.47 | < .001*** |
| GPT-4o | Llama-3.1 | 133.46 | < .001*** |
| GPT-4o | Qwen-2.5 | 249.45 | < .001*** |
| Gemini 2.5 Flash | Llama-3.1 | 197.81 | < .001*** |
| Gemini 2.5 Flash | Qwen-2.5 | 324.72 | < .001*** |
| Llama-3.1 | Qwen-2.5 | 19.73 | < .001*** |

* $p < 0.005$ ** $p < 0.001$ *** $p < 0.0005$

Table 10: Chi-square test results for pairwise Missedin-Urdu rates (Bonferroni-corrected, $\alpha = 0.005$, $N = 4{,}365$ matched instances).

| **Model A** | **Model B** | **Stat** | $p$ |
|---|---|---|---|
| Claude Sonnet 4.5 | GPT-4o | 140.29 | < .001*** |
| Claude Sonnet 4.5 | Gemini 2.5 Flash | 1.54 | 0.214 |
| Claude Sonnet 4.5 | Llama-3.1 | 0.00 | 1.000 |
| Claude Sonnet 4.5 | Qwen-2.5 | 9.42 | 0.002* |
| GPT-4o | Gemini 2.5 Flash | 0.14 | 0.706 |
| GPT-4o | Llama-3.1 | 13.08 | < .001*** |
| GPT-4o | Qwen-2.5 | 8.01 | 0.005* |
| Gemini 2.5 Flash | Llama-3.1 | 3.07 | 0.080 |
| Gemini 2.5 Flash | Qwen-2.5 | 13.60 | < .001*** |
| Llama-3.1 | Qwen-2.5 | 14.62 | < .001*** |

* $p < 0.005$ ** $p < 0.001$ *** $p < 0.0005$